\documentclass{article}
\usepackage{spconf,amsmath,graphicx}
\usepackage{algorithm}
\usepackage{algorithmic}
\usepackage{multirow}
\usepackage{amssymb}
\usepackage{subcaption}
\usepackage{booktabs}
\usepackage{bm}
\usepackage{url}
\usepackage{color}

\title{Multi-view Contrastive Learning for Online Knowledge Distillation}
\name{Chuanguang Yang$^{\star \dagger}$ \qquad Zhulin An  $^{\star \ddagger}$\sthanks{Corresponding author.} \qquad Yongjun Xu$^{\star \ddagger}$}

\address{$^{\star}$ Institute of Computing Technology (ICT), Chinese Academy of Sciences (CAS), Beijing, China \\
	$^{\dagger}$University of Chinese Academy of Sciences, Beijing, China \\
 $^{\ddagger}$ Xiamen Data Intelligence Academy of ICT, CAS, Xiamen, China\\
\{yangchuanguang, anzhulin, xyj\}@ict.ac.cn}

\begin{document}
%
\maketitle
\begin{abstract}
Previous Online Knowledge Distillation (OKD) often carries out mutually exchanging probability distributions, but neglects the useful representational knowledge. We therefore propose Multi-view Contrastive Learning (MCL) for OKD to implicitly capture correlations of feature embeddings encoded by multiple peer networks, which provide various views for understanding the input data instances. Benefiting from MCL, we can learn a more discriminative representation space for classification than previous OKD methods. Experimental results on image classification demonstrate that our MCL-OKD outperforms other state-of-the-art OKD methods by large margins without sacrificing additional inference cost. Codes are available at~\textcolor{red}{\emph{\url{https://github.com/winycg/MCL-OKD}}}.
\end{abstract}
\section{Introduction}
Albeit modern convolutional neural networks show dramatic performance in image classification tasks, it is still difficult to deploy a superb model on the resource-limited edge device. Typical solutions include efficient architecture design~\cite{zhu2019eena,yang2020gated,zhu2020efficient}, model pruning~\cite{yang2019multi,cai2020softer}, dynamic inference~\cite{hudrnet} and knowledge distillation~\cite{hinton2015distilling}. The idea of Knowledge Distillation (KD) is to transfer the useful information from an excellent yet cumbersome teacher model to a student model with low complexity. Alternatively, teacher-free \emph{Online Knowledge Distillation} (OKD)~\cite{zhang2018deep} is a more practical method for improving the performance of a given model. The framework of OKD often lies in the mutual and cooperative knowledge transfer among several student models trained from scratch.

Popular OKD methods~\cite{zhu2018knowledge,song2018collaborative,chen2019online} focus on mutually transferring instance-level class probability distributions among various student models, but neglect the more informative representational knowledge for online transfer. In this paper, we implement Multi-view Contrastive Learning (MCL) to implicitly capture correlations of encoded representations of data instances among multiple peer networks, where one peer network represents one view for understanding the input. In MCL, we try to maximize the agreement for the representations of the same input instance from various views, while pushing the representations of input instances with different labels from various views apart. Our motivation is inspired by the mechanism that various people always view a same objective in the real world with individual understandings, and their consensus is always quite robust for discriminating this objective. While human may also carry an additional inductive bias as the noise information in their understandings. Here, all peer networks build a group of people and we try to model the person-invariant  representations. 

Similar with the previous methods of OKD~\cite{zhu2018knowledge,song2018collaborative,chen2019online}, our training graph contains multiple same networks, except that additional fully-connected layers for linearly transforming representations to the contrastive embedding space, in which we perform pair-wise contrastive loss among all peer networks. Moreover, we also build an ensemble teacher from all online peer networks. The ensemble teacher transfers probabilistic knowledge to the specific student network, which is used for final deployment. Based on the above techniques, we name our framework as MCL-OKD.  

We conduct experiments on image classification tasks of CIFAR-100~\cite{krizhevsky2009learning} and ImageNet~\cite{deng2009imagenet} across widely used networks to compare MCL-OKD against other State-Of-The-Art (SOTA) OKD methods. The results show that our MCL-OKD
achieves the best performance for optimizing a given network.
Extensive experiments on few-shot classification show the superiority of MCL-OKD in metric learning for learning a discriminative feature space.

\section{Related Works}
\textbf{Contrastive learning.} The core idea of contrastive learning is to perform contrastive loss on positive pairs against negative pairs in feature embedding space. Many prior works define positive and negative pairs from two views. Deep InfoMax~\cite{hjelm2018learning} matches the input and its output from the neural network encoder. Instance Discrimination~\cite{wu2018unsupervised} learns to contrast the current embedding with previous embeddings from an online memory bank. SimCLR~\cite{chen2020simple} considers the two views of the same data sample as different augmentations, and maximizes the consistency between them. Besides two views, CMC~\cite{tian2020contrastive} and AMDIM~\cite{bachman2019learning} propose contrastive multi-view coding across the multiple sensory channels or independently-augmented copies of the input image, respectively. For multi-view learning, typical multi-view information can also be derived from multi-modal signals~\cite{smith2005development} like vision, sound and touch. In comparison, we implement MCL by leveraging multiple peer networks to encode the same data instance, which is different from creating views in terms of the data itself compared to previous contrastive learning, because our method is more advantageous to the scenario of OKD to joint training. 

\textbf{Online knowledge distillation.}
The seminal OKD method dubbed \emph{Deep Mutual Learning} (DML)~\cite{zhang2018deep} demonstrates that knowledge transfer between two peer student models during the online training achieves obviously better performance than independent training. Inspired by this insight, ONE~\cite{zhu2018knowledge} and CL-ILR~\cite{song2018collaborative} propose the frameworks which share the low-level layers to reduce the training complexity and perform knowledge transfer among various branches of high-level layers. OKDDip~\cite{chen2019online} alleviates the homogenization problem in previous ONE method by introducing two-level distillation and self-attention mechanism. All previous methods handle the probabilistic output to perform OKD, but differ in the ways of supervision. Our MCL-OKD further improves the performance of OKD from the perspective of representation learning.

\section{methodology}
\subsection{Distillation framework}
\begin{figure}[tbp]  
	\centering  
	\includegraphics[width=1\linewidth]{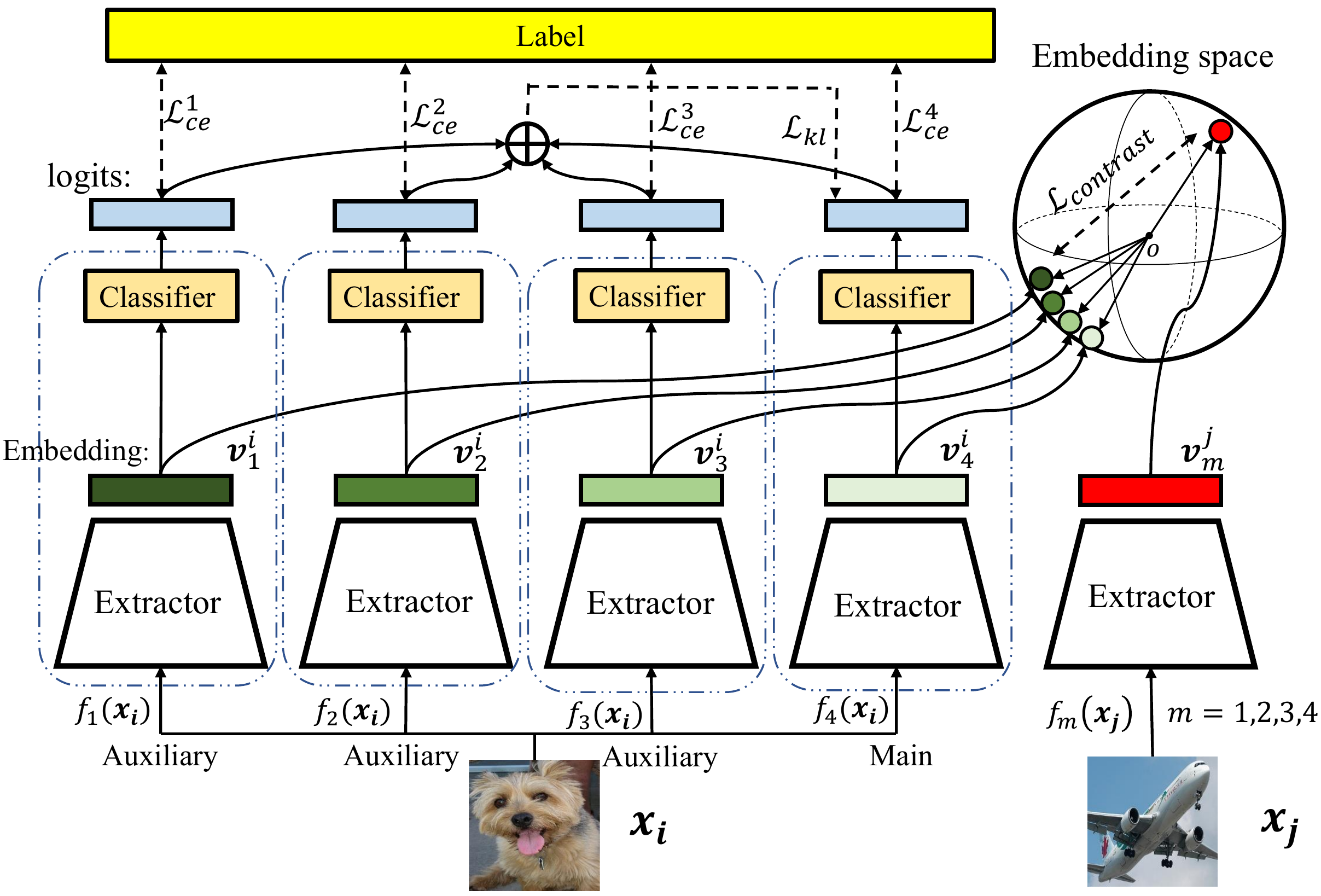}
	\caption{An overview of MCL for OKD. Given a dog, 4 networks provide 4 views for generating
		representations, contrastive learning aims to push them closed, and push the representations from instances with different classes apart.
		}  
	\label{arch}
\end{figure}

As depicted in Fig.\ref{arch}, $M$ peer networks $\{f_{m}(\bm{x})\}_{m=1}^{M}$  participate in the process of distillation during training, where each network includes a CNN feature extractor and a linear classifier. For contrastive learning, we aim to optimize the embeddings after \emph{Global Average Pooling} (GAP) layer among peer networks. We linearly transform these embeddings with $l_{2}$-normalization into the same contrastive embedding space with embedding size of 128. Given a instance $\bm{x}_{i}$, the generated contrastive embeddings of $\{f_{m}(\bm{x}_{i})\}_{m=1}^{M}$ are denoted as $\{\bm{v}_{m}^{i}\}_{m=1}^{M}$
 Similar with the previous branch-based OKD\cite{zhu2018knowledge}, low-level layers across the $M$ peer networks can also be shared to reduce the complexity and regularize the training networks.

\textbf{Training and deployment}. At the training stage, we jointly optimize  $\{f_{m}(\bm{x})\}_{m=1}^{M}$. At the test stage, we discard $M-1$ auxiliary networks $\{f_{m}(\bm{x})\}_{m=1}^{M-1}$, only the last network $f_{M}(\bm{x})$ is kept, resulting in no additional inference cost. 
\subsection{Learning objectives}
\textbf{Learning from labels.} Each network is trained by \emph{Cross-Entropy} (CE) loss between predictive probability distribution and hard labels. Given a instance $\bm{x}$ with label $y$, CE loss of the $m$-th network is:
\begin{equation}
\mathcal{L}_{ce}^{m}=-\sum_{c=1}^{C}\eta_{c,y}\log{p_{m}(y|\bm{x})}
\end{equation}
Where  $\eta_{c,y}$ is indicator that return $1$ if $c=y$ else $0$. $p_{m}(y|\bm{x})$ is the class posterior calculated from the logits distribution $[z_{m}^{1},z_{m}^{2},\cdots,z_{m}^{C}]$ by softmax normalization:
\begin{equation}
p_{m}(y|\bm{x})=\exp{(z_{m}^{y})}/[\sum_{c=1}^{C}\exp{(z_{m}^{c})}]
\end{equation}
Overall, CE loss of $M$ networks is $\mathcal{L}_{ce}=\sum_{m=1}^{M}\mathcal{L}_{ce}^{m}$.

\textbf{Distillation from an online teacher.} We simply construct an online teacher by implementing the naive ensemble for the predictive probability distribution of all networks and softened by a temperature $T$ as:
\begin{equation}
\label{teacher}
\tilde{p}_{E}(c|\bm{x})=\frac{\exp{(\frac{1}{M}\sum_{m=1}^{M}z_{m}^{c}/T)}}{\sum_{d=1}^{C}\exp{(\frac{1}{M}\sum_{m=1}^{M}z_{m}^{d}/T)}}
\end{equation}
Where $c\in \{1,2,\cdots,C\}$ and $\tilde{p}_{E}(c|\bm{x})$ denotes the soft probability of the $c$-th class. The soft probability distribution of the $M$-th network $f_{M}(\bm{x})$ is:
\begin{equation}
\label{group_leader}
\tilde{p}_{M}(c|\bm{x})=\exp{(z_{M}^{c}/T})/[\sum_{d=1}^{C}\exp{(z_{M}^{d}/T)}]
\end{equation}
We consider transferring probabilistic knowledge from online teacher to the final deployment network $f_{M}(\bm{x})$, KL divergence is thus used for aligning the soft predictions between the former and latter as:
\begin{equation}
\mathcal{L}_{kl}=\mathbf{KL}(\tilde{p}_{E}\parallel  \tilde{p}_{M} )=\sum_{c=1}^{C}\tilde{p}_{E}(c|\bm{x})\log{\frac{\tilde{p}_{E}(c|\bm{x})}{\tilde{p}_{M}(c|\bm{x})}}
\end{equation}

\textbf{Multi-view contrastive learning.} Given a training set $\mathcal{D}=\{(\bm{x}_{i},y_{i})\}_{i=1}^{N}$ includes $N$ instances with $C$ classes. We consider learning the relationship derived from different networks among various data instances: feature embeddings of the same data instance are mutually closed, while that of two data instances with different classes are far away.

Given two networks $f_{a}$ and $f_{b}$ for illustration, generated embeddings across the training set $\mathcal{D}$ are $\{\bm{v}_{a}^{i}\}_{i=1}^{N}$ and $\{\bm{v}_{b}^{i}\}_{i=1}^{N}$, respectively. We define the positive pair as $(\bm{v}_{a}^{i},\bm{v}_{b}^{i})$, and the negative pair as $(\bm{v}_{a}^{i},\bm{v}_{b}^{j}),y_{i}\neq y_{j}$. We expect to make positive and negative pairs achieve high and low cos similarities respectively by contrastive learning. Given the embedding $\bm{v}_{a}^{i}$ of instance $\bm{x}_{i}$ from the fixing view $f_{a}$, we enumerate the corresponding positive embedding $\bm{v}_{b}^{i}$ and $K$ negative embeddings $\{\bm{v}_{b}^{i,k}\}_{k=1}^{K}$ from $f_{b}$. For ease of notation, $\bm{v}_{b}^{i,k}$ is the $k$-th negative embedding relative to $\bm{v}_{a}^{i}$. We regard the optimization as correctly classifying the positive $\bm{v}_{b}^{i}$ against $K$ negative $\{\bm{v}_{b}^{i,k}\}_{k=1}^{K}$ to approximate the full distribution against all negative embeddings, which is inspired by Noise-Contrastive Estimation (NCE)~\cite{gutmann2010noise}.

The idea behind NCE-based approximation is to transform the instance-level multi-classification into  binary classification for discriminating positive pairs and negative pairs. Given the anchor $\bm{v}_{a}^{i}$, the probability of $\bm{v}_{b}^{j},j\in \{1,2,\cdots,N\}$ matching $\bm{v}_{a}^{i}$ as the positive pair is:
\begin{equation}
\label{positive_prob}
p_{p}(\bm{v}_{b}^{j}|\bm{v}_{a}^{i})=\frac{\exp{(\bm{v}_{a}^{i}\cdot \bm{v}_{b}^{j}/\tau)}}{Z_{i}},Z_{i}=\sum_{n=1}^{N}\exp{(\bm{v}_{a}^{i}\cdot \bm{v}_{b}^{n}/\tau)}
\end{equation}
Where $\tau$ is the temperature and $Z_{i}$ is a normalizing constant. Moreover, we define the uniform distribution for the probability of $\bm{v}_{b}^{j},j\in \{1,2,\cdots,N\}$ matching $\bm{v}_{a}^{i}$ as the negative pair, i.e. $p_{n}(\bm{v}_{b}^{j}|\bm{v}_{a}^{i})=1/N$. Assumed that frequency of sampling lies in every $K(K<<N)$ negative instances and 1 positive instance concurrently. Then the posterior probability of $\bm{v}_{b}^{j}$ drawn from the actual distribution of positive instance (denoted as $D=1$) is: 
\begin{align}
p(D=1|\bm{v}_{a}^{i},\bm{v}_{b}^{j})=\frac{p_{p}(\bm{v}_{b}^{j}|\bm{v}_{a}^{i})}{p_{p}(\bm{v}_{b}^{j}|\bm{v}_{a}^{i})+K\cdot p_{n}(\bm{v}_{b}^{j}|\bm{v}_{a}^{i})}
 \label{zzz}
\end{align}
We minimize negative log-likelihood derived from positive pair $(\bm{v}_{a}^{i},\bm{v}_{b}^{i})$ and $K$ negative pairs $(\bm{v}_{a}^{i},\bm{v}_{b}^{j}),\bm{v}_{b}^{j}\sim p_{n}(\cdot|\bm{v}_{a}^{i})$, which approximates the contrastive loss from $f_{a}$ to $f_{b}$ as:
\begin{align}
\label{contrastive_loss}
\mathcal{L}_{contrast}^{f_{a}\rightarrow f_{b}}=-&\log p(D=1|\bm{v}_{a}^{i},\bm{v}_{b}^{i}) \notag\\ 
+K\cdot &\mathop{\mathbb{E}}_{\bm{v}_{b}^{j}\sim p_{n}(\cdot|\bm{v}_{a}^{i})}\log[1-p(D=1|\bm{v}_{a}^{i},\bm{v}_{b}^{j})]
\end{align}
Where $\bm{v}_{b}^{j}\sim p_{n}(\cdot|\bm{v}_{a}^{i})$ denotes the randomly sampled negative embedding retrieved from an online memory bank $V$~\cite{wu2018unsupervised} instead of real-time computing along with each mini-batch, which allows us  efficiently obtaining abundant negative instances for generating contrastive knowledge. Note that only $\bm{v}_{a}^{i}$ and $\bm{v}_{b}^{i}$ are computed by $f_{a}$ and $f_{b}$ in real time from the input instance $\bm{x}_{i}$ in mini-batch.

Symmetrically, we can also fix $f_{b}$ and enumerate over $f_{a}$ by contrasting $\bm{v}_{b}^{i}$ with positive  $\bm{v}_{a}^{i}$ and negative $\{\bm{v}_{a}^{i,k}\}_{k=1}^{K}$, resulting in the contrastive loss $\mathcal{L}_{contrast}^{f_{b}\rightarrow f_{a}}$. We thus deduce the overall objective of contrastive learning between $f_{a}$ and $f_{b}$ as $\mathcal{L}(f_{a},f_{b})=\mathcal{L}_{contrast}^{f_{a}\rightarrow f_{b}}+\mathcal{L}_{contrast}^{f_{b}\rightarrow f_{a}}$. We further extend contrastive loss from two views to multiple views, which captures more robust evidences for classification as well as discards noise information. Specifically, we model fully-connected interactions among each view pair that thus lead  to $\binom{M}{2}$ relationships. Contrastive loss among $M$ networks is summarized as $\mathcal{L}_{contrast}=\sum_{1\leq a<b\leq M}\mathcal{L}(f_{a},f_{b})$.

\textbf{Overall learning objective.} We combine above three objectives to construct our final objective:
\begin{equation}
\label{overall}
\mathcal{L}_{MCL-OKD}=\mathcal{L}_{ce}+T^{2}\mathcal{L}_{kl}+\beta \mathcal{L}_{contrast}
\end{equation}
Where $T^{2}$ is used for balancing contributions between hard and soft labels, $\beta$ is a constant factor for rescaling the magnitude of contrastive loss.

\begin{table*}[t]
	\caption{Comparison of error rate (Top-1, \%) with standard deviation among OKD methods on CIFAR-100. The \textbf{bold} number denotes the lowest error rate towards row comparison with the number in brackets for showing reduction margin of error rate than second best result, which is emphasized by \underline{underline}. FLOPs denotes the number of floating-point operations to measure the computation. HCGNet achieves the best trade-off between FLOPs and accuracy across various OKD methods.}
	\label{okd}
	\centering
	\resizebox{1\linewidth}{!}{
	\begin{tabular}{cccccccc}
		\toprule
		Network&FLOPs&Baseline&DML~\cite{zhang2018deep}&CL-ILR~\cite{song2018collaborative}&ONE~\cite{zhu2018knowledge}&OKDDip~\cite{chen2019online}&MCL-OKD \\
		\midrule
		DenseNet-40-12 &0.07G&29.17$\pm $0.15&\underline{27.34}$\pm $0.36& 27.38$\pm $0.47  &29.01$\pm $0.08&28.75$\pm $0.63&\textbf{26.04}$\pm $0.25$_{(-1.30)}$ \\
		
		ResNet-32&0.07G &28.91$\pm $0.31&\underline{24.92}$\pm $0.12&25.40$\pm $0.06& 25.74$\pm $0.19&25.76$\pm $0.29&\textbf{24.52}$\pm $0.26$_{(-0.40)}$ \\
		  
		VGG-16&0.31G &25.18$\pm $0.25&24.14$\pm $0.36& \underline{23.58}$\pm $0.14&25.22$\pm $0.11&24.86$\pm $0.30&\textbf{23.11}$\pm $0.25$_{(-0.47)}$ \\
		
		ResNet-110 &0.17G&23.62$\pm $0.73&21.51$\pm $0.74&21.16$\pm $0.29&22.19$\pm $0.56 &\underline{21.05}$\pm $0.17 &\textbf{20.39}$\pm $0.59$_{(-0.66)}$ \\
		
		HCGNet-A1&0.15G &22.46$\pm $0.28&\underline{18.98}$\pm $0.20&19.04$\pm $0.17&22.30$\pm $0.57 &21.54$\pm $0.11 &\textbf{18.72}$\pm $0.21$_{(-0.26)}$ \\
		\bottomrule
	\end{tabular}}
\end{table*}
\section{Experiments}
\subsection{Dataset and setup}
	\textbf{Image Classification}. We use CIFAR-100~\cite{krizhevsky2009learning} and ImageNet~\cite{deng2009imagenet} benchmark datasets for evaluations. CIFAR-100 contains 50K training images and 10K test images with the input resolution 32$\times$32 from 100 classes, ImageNet contains 1.2 million images and 50K test images with the input resolution 224$\times$224 from 1000 classes.  We use $T=3$ and $\beta=0.025$ in equ.(\ref{overall}). Following~\cite{chen2019online}, we use $\tau=0.1$ and $\tau=0.07$ in equ.(\ref{positive_prob}) for CIFAR-100 and ImageNet respectively, and architecture-aware $K\in [256,16384]$ in equ.(\ref{contrastive_loss}). We use 4 networks in all OKD methods, \emph{i.e.} $M=4$. Given a model, training graph shares the first several stages and separates from the last two stages. Detailed experimental settings can be found in our released codes. We report the mean error rate over 3 runs.
	
	\textbf{Few-shot learning}. We use \emph{mini}ImageNet~\cite{vinyals2016matching} benchmark for few-shot classification. Prototypical network~\cite{snell2017prototypical} is used as the backbone, which also plays the role of a peer network in OKD. We use the standard data split following Snell \emph{et al.}~\cite{snell2017prototypical}. At the test stage, we report average accuracy over 600 randomly sampled episodes with 95\%
	confidence intervals for \emph{mini}ImageNet.
	For logit-based OKD methods, we add an auxiliary global classifier to the original prototypical network over the class space of training set, and perform learning of probabilistic outputs among 4 peer networks. 

\begin{table}[t]
	\caption{Comparison of error rate (Top-1, \%) on ImageNet. 'Ens' denotes the ensemble error rate. The number in brackets is the performance gain upon Baseline.}
	\label{imagenet}
	\centering
	
	\resizebox{1\linewidth}{!}{
		\begin{tabular}{ccccc}
			\toprule
			Network &Baseline&MCL-OKD&MCL-OKD (Ens) \\
			
			\midrule
			ResNet-34 & 25.43& 24.64$_{(-0.79)}$& 23.26$_{(-2.17)}$
			\\
			\bottomrule
	\end{tabular}}
\end{table}

\begin{table}[t]
	\caption{Comparison of ensemble error rate (Top-1, \%) among OKD methods on CIFAR-100.}
	
	\label{ensemble}
	\centering
	\resizebox{1\linewidth}{!}{
		\begin{tabular}{cccccc}
			\toprule
			Baseline &DML&CL\_ILR&ONE&OKDDip&MCL-OKD \\
			
			\midrule
			DenseNet-40-12 &\underline{26.02} & 26.19& 28.67& 27.51& \textbf{23.55}$_{(-2.47)}$\\
			ResNet-32 &\underline{22.97} & 24.03& 24.03& 23.73& \textbf{22.00}$_{(-0.97)}$\\
			VGG-16 &23.27&\underline{22.96}& 25.12&24.52& \textbf{22.36}$_{(-0.60)}$\\
			ResNet-110 &19.12 & \underline{18.66}&20.23&19.40& \textbf{18.29}$_{(-0.37)}$\\
			HCGNet-A1 &\underline{17.86} & 18.35& 21.64& 20.97& \textbf{17.54}$_{(-0.32)}$\\
			\bottomrule
	\end{tabular}}
\end{table}

%
%
\begin{table}[t]
	\caption{Comparison of accuracy (Top-1, \%) among KD and OKD methods on \emph{mini}ImageNet for few-shot learning.}
	\label{few-shot}
	\centering
	\begin{tabular}{lcc}
		\hline
		Method &5-Way 1-Shot&5-Way 5-Shot \\
		
		\toprule
		Baseline~\cite{snell2017prototypical} &49.10 $\pm $ 0.41 & 66.87 $\pm $ 0.33\\
		\midrule
		RKD-D~\cite{park2019relational} & 49.66 $\pm $ 0.84 &67.07 $\pm $ 0.67 \\
		RKD-DA~\cite{park2019relational} & 50.02 $\pm $ 0.83 &\underline{68.16} $\pm $ 0.67 \\
		\midrule
		CL\_ILR~\cite{song2018collaborative} &\underline{50.75} $\pm $ 0.40&67.75 $\pm $ 0.32 \\
		ONE~\cite{zhu2018knowledge} &50.67 $\pm $ 0.41 &67.58 $\pm $ 0.33 \\
		OKDDip~\cite{chen2019online}& 50.60 $\pm $ 0.42 &67.41 $\pm $ 0.33 \\
		MCL-OKD &\textbf{51.58} $\pm $ 0.41 &\textbf{69.49} $\pm $ 0.33 \\
		\bottomrule
	\end{tabular}
\end{table}

%
%

\subsection{Results of OKD methods}
\textbf{Image Classification}. Table \ref{okd} shows the comparisons of the performance among SOTA OKD methods across the various networks of VGG~\cite{simonyan2015very}, ResNet~\cite{he2016deep}, DenseNet~\cite{huang2017densely} and HCGNet~\cite{yang2020gated}. It can be obviously observed that our MCL-OKD consistently outperforms all other alternative methods of DML~\cite{zhang2018deep}, CL-ILR~\cite{song2018collaborative}, ONE~\cite{zhu2018knowledge} and OKDDip~\cite{chen2019online} by large margins, which indicates that performing MCL on the representations among various peer networks is more effective than transparent learning from probability distributions that previous OKD methods always focus on. Compared to the previous SOTA OKDDip, MCL-OKD achieves absolute 1.84\% reduction of error rate on average on CIFAR-100. Extensive experiments on the more challenge ImageNet validate that MCL-OKD significantly outperforms the baseline by 0.79\% margin.
As shown in Table~\ref{ensemble}, MCL-OKD can also achieve the best ensemble error rate among compared OKD methods with retaining all peer networks. 

\textbf{Few-shot learning}. Table \ref{few-shot} compares accuracy among SOTA KD and OKD methods on few-shot learning. We can observe that MCL-OKD significantly outperforms other OKD methods, which verifies that MCL on representations is more effective than learning on probabilistic outputs especially for metric learning tasks, due to the superiority of generating discriminative feature embeddings for instances from newly unseen classes. Moreover, MCL-OKD achieves better results than RKD~\cite{park2019relational}, which is the SOTA KD method for few-shot learning but needs a pre-trained teacher as the prerequisite.

\textbf{Training complexity.} We take ResNet-34 on ImageNet as an example. MCL-OKD uses extra 1.56 GFLOPS for contrastive computing, which is about 16\% of the original 9.60 GFLOPS by vanilla training. In practice, we observed no distinct addition of training time (1.23 hours/epoch v.s. 1.52 hours/epoch on a single NVIDIA Tesla V100 GPU). The memory bank of each peer network needs about 600MB memory for storing all 128-d features, resulting in the total 2.4GB memory for 4 peer networks.
\section{CONCLUSION}
We propose multi-view contrastive learning for OKD to learn a more powerful representation space benefiting from the mutual communications among peer networks. Experimental evidence proves the superiority of learning informative feature representations, which makes our MCL-OKD become a prior choice for model deployment in practice.

\textbf{Acknowledgements.} This work was supported by the Basic Research Reinforcement Project (2019-JCJQ-JJ-412).

\bibliographystyle{IEEEbib}
\bibliography{strings,refs}

\end{document}